# North Africa's Missing Framework: NLP-driven Mental Healthcare in Algeria and Implications for Low-resource Settings

Meriem Laifa [1,2] [0000-0002-2010-9527], Abdallah Bengueddoudj [3,4] [0009-0000-4873-6824]

[1] Department of Computer Science, Faculty of Mathematics & Computer science, University Mohamed El Bachir El Ibrahimi of Bordj Bou Arreridj, Algeria; meriem.laifa@univ-bba.dz

[2] Laboratory of Informatics and its Applications of M'sila (LIAM), M'sila, Algeria

[3] Department of Engineering, University Mohamed El Bachir El Ibrahimi of Bordj Bou Arreridj, Algeria; abdallah.bengueddoudj@univ-bba.dz

[4] Laboratory of Materials and Electronic Systems (LMSE), Bordj Bou Arreridj, Algeria

## Abstract:

Mental health disorders are a leading cause of disability worldwide, yet Natural Language Processing (NLP) research for mental healthcare has remained concentrated in high-income, English-language settings. North Africa, and Algeria in particular, is largely absent from this literature despite its unique linguistic, historical, and healthcare context. We present the first conceptual framework examining the potential role of NLP within Algeria's mental healthcare system. Drawing on narrative synthesis of global NLP mental health research, Algerian healthcare literature, and low-resource NLP methodologies, we identify four structural barriers to mental healthcare: the language-of-care gap, geographic inequities in access, stigma-related barriers to help-seeking, and the absence of research and digital infrastructure. We then map existing NLP capabilities to each barrier, outlining their potential applications, implementation constraints, and the technical, institutional, and governance requirements necessary for deployment. Based on this analysis, we propose a research and policy roadmap that prioritizes data resources, multilingual language technologies, evaluation frameworks, and regulatory capacity. Although grounded in the Algerian context, the framework addresses challenges common to many multilingual, low-resource, and post-colonial settings. This work provides a foundation for future research on culturally and linguistically appropriate NLP for mental healthcare and offers a practical roadmap for developing responsible AI-enabled mental health systems in underrepresented regions.

**Keywords:** Natural Language Processing; Digital Mental Health; North Africa; low-resource settings; post-colonial healthcare.

# 1 Introduction

Good mental health represents a state where an individual can realize their potential, manage life's normal stresses, perform productive work, and contribute meaningfully to their community [1]. It is fundamentally inseparable from and essential to overall health, encompassing the presence of psychological resilience and social functionality. However, acknowledging mental health as a fundamental component of overall health led to the unveiling of a worldwide mental health crisis that can no longer be ignored [2]. Most recently, mental health disorders became a leading cause of global disability [3], [4]. The surge in these disorders has overwhelmed healthcare systems worldwide, exposing the limitations of traditional in-person mental health services, resulting in trillions of dollars in economic losses annually [5]. However, the gap between population need and available care is widest in low-resource contexts where baseline systems are already critically under-resourced and traditional reform pathways, such as more clinicians and facilities, cannot close the distance at the required scale or speed.

A branch of Artificial Intelligence (AI) that operates directly on human language, known as Natural Language Processing (NLP), has emerged as one of the most promising technological responses to this problem. Interest in NLP applications in mental healthcare is growing at an annual rate of 19.2%, significantly outpacing the overall growth rate across scientific disciplines, which is currently at 5.08% [6], [7], [8]. This growth reflects a recognition that spoken, written, and increasingly digital language is both the primary medium through which mental health is expressed and assessed and the domain in which the most scalable, low-infrastructure interventions can be deployed [9], [10]. NLP-based tools have demonstrated measurable capacity across the full clinical spectrum: early detection and diagnosis, therapeutic intervention, continuous monitoring, and clinical workflow enhancement [8], [11], [12].

However, the geographic distribution of this research field reproduces, rather than corrects, existing global inequalities. Available NLP mental health systems are predominantly trained on English-language data (81%), with limited representation of other languages and near-total absence of the linguistic varieties spoken across the Global South [8]. While digital mental health research has expanded to sub-Saharan Africa, South Asia, and Latin America [13], [14], [15], [16], [17], [18], [19], North Africa remains underrepresented and rarely a primary focus in the NLP-based mental health literature. Algeria, the largest country in Africa [20], is largely absent from such literature and is not represented in foundational studies of digital mental health interventions, to date. This exclusion is not merely a gap in regional coverage. It reflects a systematic failure to engage with a context whose specific conditions generate analytical and methodological insights that more researched settings cannot provide.

Algeria's case is not simply one instance of a general developing country problem. It represents a convergence of four conditions that, taken individually, appear in other African or Middle East and North Africa (MENA) contexts, but no comparable country presents them simultaneously. First, Algeria's psychiatric system was structurally shaped by French colonial governance and maintained French mental health law until 1975, relatively longer than any comparable postcolonial state, producing a language-of-care infrastructure in which clinicians are trained in French while the majority of patients communicate in Algerian Arabic dialects or Berber languages [21], [22]. Second, Algeria's multilingual landscape has minimal digital corpora and is absent from existing NLP systems [23]. Third, the 1990s civil conflict, Algeria's so-called "dark decade", produced a trauma-burdened population [24]. This epidemiological profile diverges sharply from the populations on which existing NLP mental health models have been trained. Fourth, Algeria has produced no published peer-reviewed research on NLP in mental healthcare, no benchmark clinical datasets in local languages, and no validated digital mental health tools adapted to its context. While Algeria has made measurable progress in broader digital infrastructure, its mental healthcare system has yet to begin a meaningful digital transition, which is a gap that makes our paper both timely and necessary.

The convergence of the above-mentioned conditions matters not only for mental healthcare generally but for NLP methodology specifically, and the relationship between the two is constitutive. Algeria's mental healthcare system is failing across several simultaneous dimensions that NLP is uniquely positioned to address, not because it is the most advanced available technology, but because it is one of the few intervention solutions that can function without the physical infrastructure, clinical workforce density, or pre-existing digital health ecosystems that other solutions presuppose. The language-of-care gap demands tools that operate across code-switching and dialect variation that no current system supports. The scale of geographic exclusion - where 96.13% of psychiatric resources serve the northern 56% of the population [25] - requires interventions that function without consistent clinical availability or physical access. On the other hand, the severity of stigma makes the social visibility of help-seeking itself a clinical barrier, rendering anonymous text-based interaction not a convenience but a clinical requirement for reaching populations who will not present for formal services. Moreover, the near-total absence of digitized clinical data means that properly designed NLP tools could simultaneously deliver care and generate the data infrastructure that evidence-based practice will eventually require.

None of these conditions uniquely distinguishes Algeria, but their co-occurrence does. It is precisely this convergence that makes NLP not merely applicable here but structurally necessary, and that defines what an NLP system for this context would need to look like and why building it requires a research and policy agenda that does not currently exist.

Recent scholarship has begun to address adjacent territories. Alasmari [26] conducted a scoping review of Arabic NLP techniques applied to mental health across Arabic-speaking populations, documenting the predominance of transformer-based models for depression and suicidality detection from social media data. Mikaeili et al. [27] proposed the MENA-MHPI model, a culturally adapted framework for mental health promotion across the Middle East, combining digital innovation with community-based psychological support. On the other hand, Naeim et al. [28] provided a bibliometric mapping of twenty-five years of mental health research across the Middle East, which is an analysis that strikingly does not mention Algeria or North Africa, confirming through its omission the exclusion of this subregion from mainstream MENA mental health scholarship. Taken together, these contributions establish that regional interest in digital mental health is real and growing, but that no existing work has constructed a framework examining NLP applications within a single North African country's healthcare system, analyzed at the intersection of its specific linguistic complexity, colonial infrastructure legacy, trauma epidemiology, and absent research infrastructure.

In this perspective paper, we present the first original framework examining how NLP can be applied to mental healthcare in a multilingual, low-resource, post-colonial North African context, using Algeria as an in-depth case study. Rather than reviewing existing NLP applications, we interrogate their relevance against four systematic "failure axes" that structure mental healthcare access in this context: the language-of-care gap between patients and formal services; geographic exclusion from care provision; stigma-driven barriers to help-seeking; and the absence of locally grounded research and digital infrastructure. These axes delineate the conditions under which current NLP systems, which largely develop in high-resource environments, fail to translate effectively. Our contribution is to articulate how these constraints redefine the technical, institutional, and governance requirements for NLP in such settings. Although grounded in Algeria, the framework is explicitly designed for transferability: it operates as a structured mapping between recurrent systemic constraints and the corresponding adaptations required in system design, deployment, and policy. Variants of this four-axis configuration recur across much of the post-colonial Global South, supporting the framework's applicability beyond the Algerian context.

This paper advances a perspective grounded in an evidence-informed synthesis rather than a systematic or scoping review. It integrates insights across heterogeneous sources to construct a conceptual and policy-relevant framework, rather than exhaustively survey the literature. Evidence was identified through targeted searches of PubMed, Google Scholar, and Scopus, complemented by grey literature including Algerian government health reports, World Health Organization (WHO) country profiles, and regional policy documents. In the absence of a dedicated evidence base on NLP and mental health in Algeria, the analysis draws on three intersecting domains: global research on NLP in mental health, documentation of the Algerian mental health system, and methodological advances in low-resource NLP. Their synthesis enables the formulation of a framework not contained within any single literature. The scope is restricted to NLP-based interventions; broader digital mental health tools are considered only insofar as they contextualize language-based approaches.

The paper proceeds as follows. Section 2 offers a contextual overview of mental health in Algeria, examining the historical foundations of its healthcare system, the current state of its infrastructure and resources, and the societal and cultural factors that shape help-seeking and treatments. Section 3 addresses the necessity of NLP as an intervention in Algerian mental healthcare, followed by Section 4 which maps NLP capabilities onto Algeria's four structural failure axes, arguing that for each axis, NLP can address specific barriers. Section 5 provides a structured research and policy roadmap organized by time horizon. Finally, Section 6 concludes the paper with a summary of the main takeaways and a discussion of the framework's potential applicability to comparable low-resource settings.

# 2 Contextual Overview of Mental Health in Algeria

Comprehensive epidemiological data on mental health in Algeria remain fragmented and limited, a pattern consistent across the MENA region [29], [30] . Available evidence indicates that mental disorders affect 10-15% of the Algerian population, with depression (5-7%), anxiety (4-6%), Post Traumatic Stress Disorder (PTSD), and substance use disorders among the prevalent conditions [23], [31], [32]. Population-specific studies suggest these figures are conservative: anxiety disorder prevalence reaches 43% in Algiers (the capital), with 13% presenting PTSD [33]. On the other hand, reported suicide rates fall below global averages (2.5 per 100,000) [34], though these figures are widely considered unreliable given the cultural stigma attached to public disclosure of psychological distress. The absence of a standardized national surveillance system itself constitutes a structural failure with direct implications for NLP research: there is no data infrastructure on which computational tools could be trained or validated.

## 2.1 History

Algeria's psychiatric system was structurally constituted by French colonial governance and has never fully shed that constitution. Following the 1830 invasion, patients with severe disorders were transferred to asylums in southern France [35]. Formal mental health legislation was not extended to Algeria until 1878, and when it was, it applied the French law of 1838 without adaptation to local cultural or linguistic realities [21], [36]. By independence in 1962, Algeria had approximately 6,000 psychiatric beds built within a francophone institutional framework in which clinical training, documentation, and patient interaction were conducted in French, while the majority of the population communicated in Arabic dialects or Berber languages [21], [37].

Critically, this framework did not end with independence. Algeria continued to operate under the French mental health law into 1975, until the introduction of the second national mental health law in 1985, which remains in force despite sustained criticism of its implementation [22]. A third legislative attempt embedded within the 2018 national health law has not resolved the system's fundamental resource and governance failures [23], [38]. The institutional consequence is a psychiatric infrastructure whose language of care remains structurally misaligned with the language of its patients. This misalignment has not been addressed by any recent legislative reform, but NLP is uniquely positioned to begin closing it.

## 2.2 Current State of Mental Healthcare Infrastructure and Resources in Algeria

Despite a fourfold population increase since independence, Algeria's psychiatric bed count has fallen from approximately 6,000 in 1962 to 5,299 today, distributed across 19 psychiatric hospitals, 27 psychiatry services in general hospitals, and six university hospital units [23]. Outpatient infrastructure that contains 161 intermediate mental health centers, 270 private clinics, and 303 psychology offices is insufficient and geographically concentrated, resulting in overcrowded inpatient wards and minimal community-based follow-up care [23], [33].

The geographic distribution of these resources constitutes a structural exclusion rather than an access gap. The northern region, housing 56.2% of the population, holds 96.13% of psychiatric facilities; the Hauts Plateaux (31.2% of population) and the South (containing 12.4%) together hold only 3.87% of psychiatric facilities [25]. For the majority of Algeria's territory, geographic location is a more decisive determinant of mental health access than clinical need or economic status. This is the condition that NLP-based remote monitoring and mobile-first interventions are specifically positioned to address.

Furthermore, Algeria has approximately 2.27 psychiatrists per 100,000 people [32], that is far below the recommended 1:10,000 ratio [39]. On the other hand, specialists' emigration to higher-income countries and private practice is accelerating existing workforce shortages [40]. With over 81% of a limited mental health budget directed to inpatient care, community-based and preventive services remain structurally underfunded [33]. This workforce and infrastructure deficit cannot be closed at scale through conventional means, and it is precisely the structural condition that makes scalable, low-infrastructure NLP interventions not merely useful but necessary.

## 2.3 Algerian Societal Factors and Cultural Attitudes Toward Mental Health

Stigma in Algeria operates not merely as a social discomfort but as a clinical barrier with structural consequences. Mental illness is widely framed as personal weakness or familial dishonor, leading families to conceal rather than treat affected members, especially women [23], [41]. This dynamic systematically suppresses help-seeking and produces underreporting, meaning that even the limited epidemiological data available likely understates true prevalence.

Moreover, Algeria's trauma burden is epidemiologically distinctive and directly relevant to NLP system design. The 1990s civil war conflict (aka the Dark Decade) produced documented rates of 39.5% PTSD, 23.3% major depression, and 38% anxiety disorders among adults, with 91.9% reporting traumatic event exposure [24]. This population's symptom profile, linguistic expression of distress, and help-seeking behaviors diverge sharply from those of Western populations on which existing NLP models have been trained. For instance, a system trained on Reddit depression posts or English clinical transcripts will not generalize to this context.

Cultural and structural barriers compound each other in ways specific to the Algerian context. Traditional healing frameworks, particularly *Rokia* (i.e., Quranic healing) and the concept of *Junoon*, (i.e.,in which mental illness is attributed to jinn possession) position distress as spiritual rather than clinical, thereby creating active competition with psychiatric intervention [21], [33]. Simultaneously, language barriers fracture the clinical encounter itself: psychiatrists trained in French frequently cannot conduct interviews in the Algerian Arabic dialects or Berber languages that patients use to express psychological distress [42], [43]. Furthermore, cultural distrust of formal institutions reinforces avoidance of in-person services, as hospitals had been culturally associated with dying rather than healing [44]. Together, these conditions make anonymous, culturally-adapted, linguistically-appropriate digital interaction not a technological convenience but a requirement for reaching populations that formal services fail to reach.

# 3 NLP as a Necessary Intervention

Multiple scalable digital health modalities have demonstrated measurable impact in low-resource settings, and the case for NLP requires distinguishing it from these alternatives on grounds specific to Algeria's conditions rather than on general promises of scalability or cost-effectiveness.

The first advantage concerns NLP's infrastructure floor. Telepsychiatry requires stable video connectivity and a clinician available at both ends of the call, neither of which can be guaranteed across Algeria's Hauts Plateaux and southern regions, where the psychiatrist-to-population ratio approaches zero. Additionally, task-shifting to community health workers requires training infrastructure and a supervisory workforce that does not exist at the required scale. Mental health teletherapy platforms presuppose digital literacy, existing data ecosystems, and technical support capacity that Algeria's healthcare system cannot currently provide. NLP-based interventions operate at a lower infrastructure floor than any of these: a text message exchange, a voice recording, or a dialect-aware chatbot conducting a check-in in a local language [11], [12]. This is not an argument that NLP is superior to the alternatives in all contexts. It is an argument that NLP is the intervention class whose minimum operating requirements Algeria can currently meet.

On the other hand, properly designed NLP tools can simultaneously deliver care and generate the annotated clinical data that evidence-based practice will eventually require — addressing Algeria's absent research infrastructure as a byproduct of deployment rather than as a separate program requiring separate funding and separate timelines [8], [12]. No other available intervention class offers this dual function. The following section operationalizes both advantages by mapping NLP capabilities onto each of Algeria's four failure axes specifically.

# 4 NLP Applied to Algeria's Four Failure Axes: a Framework

The following framework maps NLP capabilities onto the four structural conditions established in Section 2. For each axis, we identify the specific barriers that define the problem in the Algerian context, the NLP capabilities positioned to address them, and the implementation requirements that do not yet exist. Algeria's digital infrastructure expansion, reflected in its advancement of 15 positions in the ITU Development Index to rank 74th globally, with over 5.8 million households connected and a national fiber-optic network exceeding 200,000 kilometers [45], [46], [47] , indicates that mobile-first NLP interventions are now deployable at a scale that was not feasible a decade ago. The infrastructure precondition for the interventions described below is therefore increasingly satisfied at the national level, even if geographic disparities within the country persist. The four axes are summarized in **Table 1**, along with the core barriers, intervention potential, and foundational implementation requirements.

## 4.1 The Language-of-Care Gap

Algeria's psychiatric system conducts its clinical work mainly in French, the language of its training infrastructure, while the majority of patients communicate in Algerian Arabic dialects, Kabyle, Chaoui, Nozabite, Tamahaq, or other Berber varieties that lack standardized written forms and are absent from existing NLP systems [42]. This is not a translation problem. It is a structural misalignment between the language of institutional medicine and the language in which psychological distress is experienced and expressed, which can have significant consequences on diagnosis. Symptoms' presentation is filtered through imperfect code-switching when clinical assessments are conducted in the patient's second or third language. Additionally, the linguistic markers most relevant to mental health detection (e.g., affect-laden vocabulary, idioms of distress, and culturally-specific metaphors) are precisely those most likely to be suppressed or distorted when a patient cannot speak in their primary language.

NLP is positioned to address this axis at two levels. At the clinical encounter level, real-time interpretation and dialect-aware transcription tools could bridge the language gap between clinicians and patients, enabling more accurate symptom elicitation and reducing the diagnostic distortion produced by language mismatch. Combining NLP with Optical Character Recognition (OCR) could additionally unlock the clinical intelligence embedded in Algeria's extensive paper-based records, the majority of which are written in French, thereby transforming previously inaccessible archival data into analyzable clinical evidence [48]. At the population level, NLP models capable of processing Algerian Arabic dialects and Berber varieties could enable passive screening through social media monitoring and active screening through chatbot-based assessment instruments, reaching individuals who would not present to French-language clinical services.

The barriers at this axis are foundational and cannot be resolved by deploying existing tools. Most NLP-based mental health systems are trained on English-language data (81%), with Modern Standard Arabic (MSA) representing only 1.5% of training corpora, and Algerian Arabic dialects and Berber varieties representing effectively zero [8]. Algerian dialects differ markedly from MSA and from each other: geographic isolation has produced distinct dialectal communities, urban contact zones have generated hybrid forms, and centuries of Arabic-Berber-French language contact have created code-switching patterns that no existing monolingual or bilingual NLP model can process [23]. Berber varieties present an additional challenge; several lack standardized orthographies, meaning that annotated text corpora cannot be built without first resolving contested standardization questions. Addressing this axis requires building annotated datasets in Algerian Arabic dialects and, at a minimum, the major Berber varieties from the ground up, in collaboration with linguists, clinical psychologists and psychiatrists, and community representatives. No shortcut through existing Arabic NLP resources is available; the linguistic distance is too great and the clinical stakes of misclassification is too high.

At the level of deployment, these constraints translate into specific system requirements: dialect-aware models capable of handling non-standardized and code-switched input; clinical tools that preserve, rather than normalize away, culturally embedded expressions of distress; and infrastructures that integrate both oral language processing and legacy French-language records, for instance, through coupling NLP with OCR. At the population level, they necessitate models capable of operating in informal, dialect-rich environments, enabling screening pathways that do not depend on French-language clinical access.

## 4.2 Geographic Exclusion

With 96.13% of psychiatric resources concentrated in the northern region serving 56.2% of the population, the Hauts Plateaux and the South—together containing 43.8% of Algeria's territory and approximately 43.6% of its population—have effectively no formal mental healthcare infrastructure [25]. For these populations, the question is not how to improve mental healthcare but how to provide any at all. Conventional scaling strategies, such as training more clinicians or building more facilities, cannot close this gap at the required speed or scale. The structural condition demands interventions that do not require a clinician's presence or physical access.

NLP-based tools offer the most viable pathway to reaching these populations. Mobile-based chatbots capable of conducting mental health check-ins in local dialects can provide continuous monitoring for patients unable to access regular care due to distance, without requiring clinician availability [11]. Wearable and mobile tools equipped with NLP capabilities can detect early warning signs of relapse in conditions such as schizophrenia or bipolar disorder and relay alerts to distant clinicians, enabling timely intervention despite geographic separation [48]. NLP tools can summarize patient histories from audio inputs in different languages, providing asynchronous clinical documentation support and reducing the administrative burden on the small number of clinicians serving dispersed populations, thereby extending their effective capacity. For Algeria's rural and southern populations specifically, these are not enhancements to existing care; they represent the only scalable pathway to any care.

The barriers at this axis are divided between technical and structural. Technically, connectivity in the Hauts Plateaux and southern regions, while improving, remains inconsistent. Consequently, NLP tools designed for this context must be optimized for low-bandwidth environments and capable of offline functionality. The usability of tools designed for urban, educated users is also limited by digital literacy among older, rural, and economically marginalized populations. Structurally, the absence of Electronic Health Record (EHR) systems in most facilities outside major cities means there is no digital clinical record infrastructure into which NLP tools can integrate. Algeria's EHR pilot, launched in 2012 at three urban sites without a unified national patient identifier, remains geographically and institutionally limited [49]. NLP deployment at this axis, therefore, cannot wait for EHR infrastructure to mature; it requires standalone tools designed to function in paper-record environments and generate their own data trails.

## 4.3 Stigma-Driven Barriers to Help-Seeking

Mental illness framed as familial dishonor, the cultural association of hospitals with dying, and the institutional distrust embedded in Algerian social norms collectively ensure that significant proportions of the population experiencing mental distress will not present to formal services regardless of geographic access or financial cost [21], [23], [44]. The prevalence of traditional healing frameworks positions clinical intervention as culturally competing rather than culturally continuous with existing help-seeking [33]. For these populations, the social visibility of help-seeking is itself a clinical barrier. Anonymity is not a feature but a condition of access in many cases.

NLP-based conversational agents uniquely position themselves at this axis because they can provide a private, stigma-free interaction space that formal services structurally cannot. Text-based and voice-based chatbots allow individuals to disclose distress without the social risk of being seen at a clinic or identified as seeking mental healthcare [11]. Multiple chatbot-based systems have demonstrated measurable reductions in depression, anxiety, and loneliness symptoms across general and frontline worker populations, including in low-resource settings [11], [50]. Beyond individual interaction, NLP tools can monitor social media platforms and community digital spaces for signals of psychological distress, enabling population-level screening that reaches individuals before they would ever self-identify as needing care. In a context where cultural norms suppress active help-seeking, this passive screening function proves particularly valuable. Additionally, NLP systems analyzing public discourse can identify and target mental health misinformation, including stigmatizing language and supernatural illness attribution, and enable precision psychoeducation campaigns calibrated to specific community contexts.

The barriers at this axis are primarily cultural and methodological rather than infrastructural. Cultural adaptation is a precondition, not an enhancement. For example, NLP tools that ignore the Junoon framework, present clinical language that conflicts with Islamic therapeutic frameworks, or operate in formal Arabic that patients do not use for emotional expression, will not be trusted and will not be used. This requires not just translation but the co-design of tools with clinicians, cultural consultants, and community members from the populations they are intended to serve. In a context devoid of formal clinical alternatives, the risk of emotional dependency on AI-based conversational agents increases methodologically. Users who form therapeutic relationships with chatbots in the absence of referral pathways face a clinical risk that is ethically distinct from the same risk in high-resource settings where human follow-up is available [48]. Governance frameworks that define the boundaries of chatbot intervention and mandate human escalation protocols are, therefore, not optional safeguards but clinical necessities specific to this axis.

## 4.4 Absent Research and Digital Infrastructure

Algeria has not produced enough published NLP mental health research in peer-reviewed journals. The earliest documented work in this domain consists of three studies appearing only in 2024—two conference papers investigating sentiment analysis and transfer learning for chronic disease patients' mental health support [51], [52], and one journal paper applying a CNN-based NLP method to psychiatric disorder diagnosis across existing datasets [53]. This is not a research gap but a research absence, and it compounds every other failure axis: tools cannot be built without data, data cannot be collected without governance frameworks, and governance frameworks cannot be designed without institutional capacity that the brain drain of specialists continues to erode.

NLP's most important structural advantage at this axis is its dual function: properly designed NLP tools can simultaneously deliver care and generate the annotated data that evidence-based practice will eventually require. A dialect-aware chatbot deployed for mental health screening not only screens, but it also produces transcripts that, with appropriate consent and privacy protocols, constitute a corpus of mental health-related language in Algerian dialects. Additionally, a clinical documentation support tool generates structured records from unstructured clinical encounters. This data-generative function means that NLP deployment, if designed with research infrastructure in mind from the outset, can begin closing the data gap as a byproduct of care delivery rather than as a separate research program requiring separate funding and separate timelines. Privacy-preserving methodologies (e.g., federated learning, differential privacy, secure multi-party computation, and encrypted data enclaves) make this possible without centralizing sensitive patient data, addressing the confidentiality concerns that constitute a legitimate barrier to data collection in mental health contexts [12], [48].

The barriers at this axis are institutional and governance-related. Algeria currently has no public regulatory framework for AI in healthcare, no established guidelines for NLP-based system validation in clinical contexts, and no national data governance policy for mental health information. The regulatory vacuum creates liability and uncertainty that discourages institutional investment and makes it difficult for researchers to design studies with clear ethical parameters. Addressing this axis requires parallel tracks: technical work to build the first annotated datasets and pilot tools and policy work to create the governance infrastructure within which larger-scale deployment can proceed responsibly. However, these tracks cannot be sequential; waiting for governance frameworks before beginning technical work will delay the research program by years in a context where the clinical need is immediate.

***Table 1.*** *Algeria's mental healthcare failure axes: core barriers, intervention potential, and foundational implementation requirements*

| Failure Axis | Core challenges in the Algerian Context | NLP Capabilities | Primary Implementation Requirements |
|---|---|---|---|
| ***Language-of-care gap*** | Clinicians trained in French; patients communicate in dialects and Berber varieties absent from existing NLP systems | Dialect-aware screening; real-time interpretation; OCR-NLP for paper records | Annotated corpora in Algerian Arabic dialects and Berber varieties; linguist-clinician-AI collaboration |
| ***Geographic exclusion*** | 96.13% of resources serve the northern 56% of the population; not enough clinical workforce in the southern regions | Mobile chatbots for remote check-ins, relapse detection, asynchronous documentation support | Low-bandwidth tool design; standalone deployment without EHR dependency |
| ***Stigma-driven barriers*** | Help-seeking is socially visible; institutional distrust; supernatural illness frameworks compete with clinical intervention | Anonymous conversational agents; passive social media screening; targeted psychoeducation | Cultural co-design; human escalation protocols; governance frameworks defining chatbot limits |
| ***Absent research infrastructure*** | No peer-reviewed NLP mental health research; no annotated datasets; no data governance framework | Data-generative deployment; privacy-preserving data collection; federated learning | Parallel technical and policy tracks; institutional capacity building; data governance legislation |

# 5 Research Directions and Recommendations

Translating the framework presented in Section 4 into practice requires a research and policy agenda that acknowledges Algeria's current starting position. The following roadmap organizes priorities by time horizon, with each recommendation linked to the primary failure axis it addresses. Two principles apply across all phases. First, NLP tools should function as supplements to clinical judgment rather than replacements for it; in a system already critically short of clinicians, tools that displace rather than extend human capacity would exacerbate the workforce crisis they are intended to mitigate [7]. Second, governance infrastructure must be developed in parallel with technical advancement rather than sequentially, as delaying research until regulatory frameworks are fully established would significantly slow progress in a context where clinical need is immediate [48]. **Figure 1** summarizes the proposed research and policy roadmap.

## 5.1 Short-Term Priorities (0-2 years)

The immediate priority is building the foundational infrastructure without which no subsequent technical work is possible. Four actions are required at this phase.

### Dataset construction:

The first and most urgent technical priority is the creation of annotated mental health datasets in Algerian Arabic dialects and the major Berber varieties. No NLP tool for this context can be built, validated, or deployed without this foundation. Dataset construction requires structured collaboration between computational linguists, mental health clinicians and specialists, and community representatives, with privacy-preserving collection protocols, including federated learning and encrypted data enclaves embedded from the design stage rather than added retrospectively [12]. Ethical standards and annotation protocols must be established before data collection begins, not after.

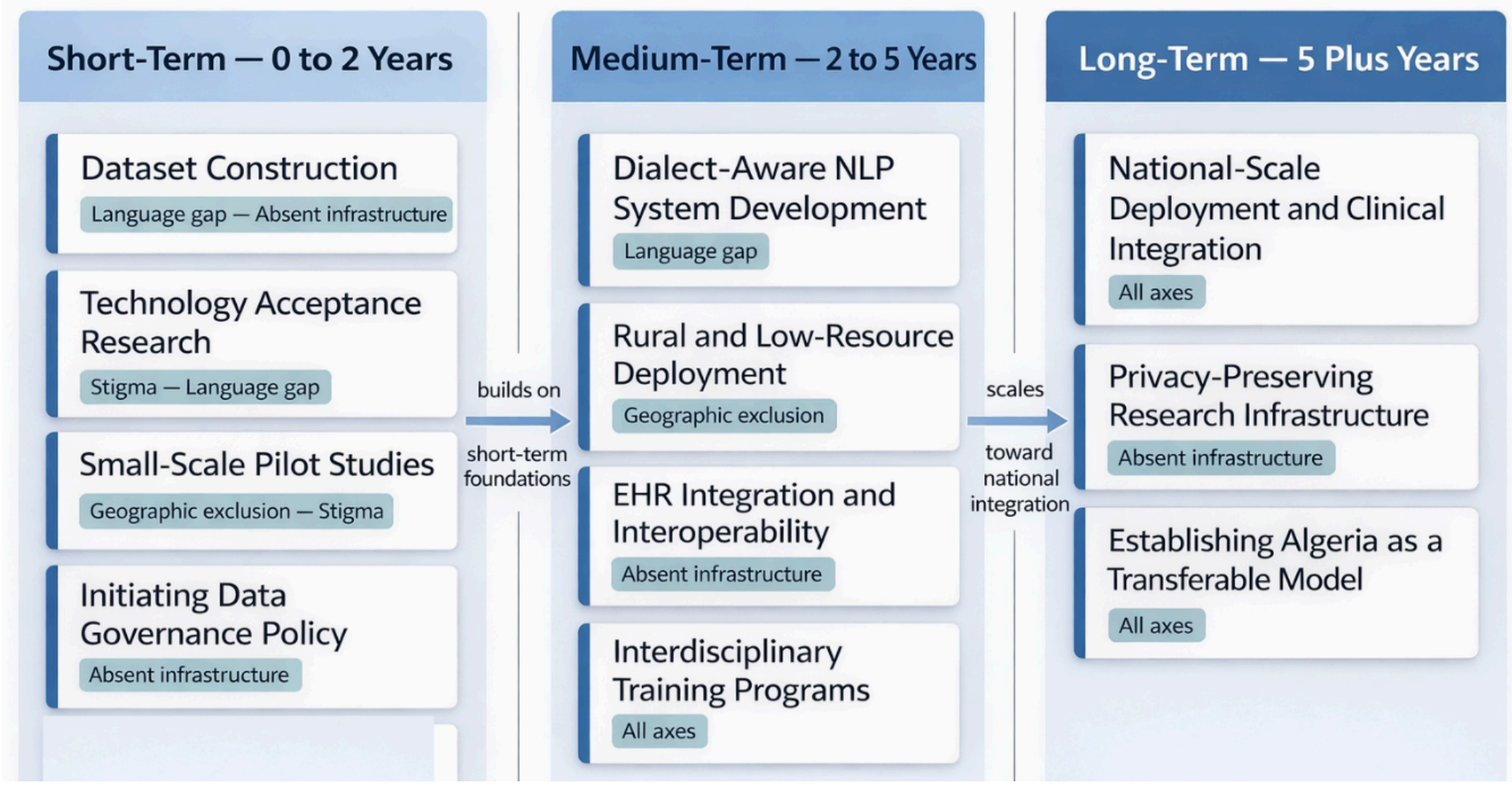

***Figure 1.*** *NLP research and policy roadmap for Algeria*

### Technology acceptance research:

Even a technically sound NLP tool is clinically useless if it is not accepted by the populations it is designed to serve. Algeria-specific technology acceptance studies that examine both clinician and patient attitudes toward NLP-assisted mental health tools are a prerequisite for deployment design [54]. This research must account for the specific trust dynamics of the Algerian context: institutional distrust of formal healthcare, cultural frameworks that position mental illness as spiritual rather than clinical, and the social risk of being identified as a mental health technology user. Existing technology acceptance models developed in Western or high-resource contexts require adaptation before application here.

### Small-scale pilot studies:

Early implementation should begin with tightly scoped pilot studies in urban mental health clinics where digital infrastructure and clinical oversight are available. Pilots should target conditions with well-defined diagnostic criteria to enable performance evaluation, and should assess not only technical accuracy but clinical utility, user experience, and unintended consequences [11]. The evidence generated by these pilots provides the empirical foundation for expansion and the data required to make the investment case to policymakers operating under constrained budgets.

### Initiating data governance policy:

Regulatory frameworks for AI in healthcare cannot wait for technical tools to be deployment-ready. Policymakers, specifically the ministries of health, higher education, and digital transformation, must begin developing national guidelines for NLP-based system validation, clinical integration standards, and mental health data governance in parallel with the technical work described above. The regulatory vacuum currently discourages institutional investment and creates ethical uncertainty for researchers. Initiating this process now, even if frameworks are not finalized until the medium term, compresses the overall timeline.

## 5.2 Medium-Term Directions (2-5 years)

With foundational datasets, initial pilot evidence, and emerging governance frameworks in place, the medium term focuses on building advanced systems and extending deployment beyond urban clinical settings.

**Dialect-aware and multilingual NLP system development:**

Drawing on the datasets constructed in the short term, researchers should develop NLP systems capable of navigating Algeria's full multilingual landscape, including code-switching between Arabic dialects, Berber varieties, French, and English within single patient interactions. This is technically demanding work that requires sustained funding and computational linguistics expertise not currently present in Algerian institutions at scale. International research partnerships are essential at this phase: structured collaborations that transfer knowledge to local teams while building domestic capacity, rather than extracting data and publishing results from outside. Dialect-aware systems should be validated against clinical outcomes, not only against benchmark NLP datasets, to ensure that technical performance translates to clinical relevance.

**Rural and low-resource deployment:**

Evidence from urban pilots should inform the design of tools specifically optimized for low-bandwidth environments and users with low digital literacy, who constitute the majority of geographically excluded populations in Algeria. Mobile-first chatbot tools for remote mental health check-ins, asynchronous clinical documentation support, and NLP-enabled relapse monitoring should be tested in Hauts Plateaux and southern region settings, where the absence of clinical infrastructure makes NLP not an enhancement but the primary available intervention. At this phase, the tool design must prioritize offline functionality and assume that no EHR integration exists in these settings.

**EHR integration and interoperability:**

Algeria's EHR pilot, stalled since its 2012 launch at three urban sites [49], must be advanced in parallel with NLP tool development, not as a prerequisite to it, but as a medium-term integration target. NLP tools deployed in the short term should be designed with future EHR interoperability in mind, generating structured data outputs that can be absorbed into national health information systems as those systems mature. The interoperability standards established at this phase will determine whether Algeria's NLP mental health infrastructure scales coherently or fragments into disconnected local deployments.

**Interdisciplinary training programs:**

The workforce gap in NLP-informed mental healthcare operates in both directions: clinicians lack digital literacy and NLP tool competency, while data scientists lack clinical knowledge of mental health assessment and therapeutic principles. Medium-term training programs must address both sides simultaneously—equipping psychiatrists, psychologists, and social workers with NLP tool utilization and AI ethics while training computational researchers in the clinical realities of mental health assessment, cultural competency, and the specific dynamics of the Algerian context. Universities are the natural institutional home for these programs and should be prioritized as sites of interdisciplinary NLP-mental healthcare research.

### 5.3 Long-Term Goals (5+ years)

The long-term phase assumes functioning annotated datasets, validated tools, trained interdisciplinary workforces, and established governance frameworks. Its focus is national integration, evidence consolidation, and establishing Algeria as a transferable model for comparable settings.

**National-scale deployment and clinical integration:**

Validated NLP tools should be integrated into Algeria's national mental health program as standard components of care delivery and not as experimental additions, with ongoing monitoring of clinical outcomes, equity of access across geographic regions, and unintended consequences. Integration at this scale requires sustained government commitment, dedicated mental healthcare technology funding within the national health budget, and clinical leadership that understands both the capabilities and the limits of NLP-assisted care.

**Privacy-preserving research infrastructure:**

As deployment scales, the volume of mental health data generated creates both research opportunities and confidentiality obligations. Long-term research infrastructure should implement secure multi-party computation and differential privacy methods that allow collaborative model training across institutions without centralizing sensitive patient data [12], [48]. This infrastructure positions Algeria to participate in international NLP-based mental health research collaborations as a data-contributing partner rather than a recipient of externally developed tools.

**Establishing Algeria as a transferable model:**

The four-condition convergence analyzed in this paper, namely colonial infrastructure legacy, multilingual complexity, trauma-specific epidemiology, and absent research infrastructure, recurs across much of the post-colonial Global South in varying configurations. The framework developed here, and the implementation evidence generated through Algeria's research program, should be documented and published in forms accessible to researchers and policymakers in comparable settings. Algeria's value to the global NLP mental health literature is not only what it gains from international knowledge. It is what it generates through the process of building a research and clinical infrastructure where none previously existed.

## 6 Conclusion

Algeria’s mental healthcare system is not failing because of a single correctable deficiency. It is failing across four simultaneous and interconnected conditions: a colonial infrastructure legacy that misaligned the language of care from the language of patients, a geographic distribution of resources that excludes the majority of the country’s territory, stigma dynamics that make the social visibility of help-seeking a clinical barrier in itself, and a complete absence of research and data infrastructure on which evidence-based intervention depends. What this paper has argued, and what the presented framework established, is that these four conditions share a common domain, namely language, and that NLP, as the computational discipline that operates directly on human language in context, is therefore not merely applicable here but necessary. This is a precise claim, not an optimistic one. It does not assert that the NLP will solve Algeria’s mental health crisis. It asserts that NLP is one of the few intervention classes capable of functioning within the specific constraints this crisis has produced, and that building it for this context requires a research and policy agenda that does not currently exist.

Such an agenda is not an option, and it is not premature. The framework maps what must be built and organizes these requirements by time horizon. The roadmap is designed to be actionable by researchers, clinicians, and policymakers operating under real constraints of Algeria’s current institutional environment, not as an aspirational vision contingent on resources that do not exist but as a structured sequence of steps that begins with what is possible now.

Whether the four-condition convergence analyzed in this paper recurs with sufficient consistency across other post-colonial low-resource settings to make this framework directly transferable remains an empirical question that this paper cannot solve. What can be asserted is that the analytical structure developed here offers a methodological approach applicable beyond the Algerian case. Future research testing this framework in comparable settings will determine its generalizability.

The primary contribution of this paper is the construction of an analytical foundation that did not previously exist: the first framework examining NLP applications within a North African mental healthcare system, grounded in the specific linguistic, historical, and institutional conditions of Algeria rather than adapted from frameworks designed for better-resourced contexts. The value of this foundation lies not only in what it describes for Algeria but also in what it demonstrates methodologically: namely, that productive NLP mental health research is possible in contexts of near-total data absence, provided that the framework is built from the structural realities of the context rather than from the assumptions of the literature that preceded it.